\documentclass{article}
\usepackage{spconf,amsmath,epsfig}
\usepackage{subcaption}

\usepackage{color,xcolor}
\usepackage[pagebackref=true,breaklinks=true,letterpaper=true,colorlinks,bookmarks=false]{hyperref}

\newcommand{\todo}[1]{}
\renewcommand{\todo}[1]{{\color{red} TODO: {#1}}}

\newcommand{\secref}[1]{Section~\ref{sec:#1}}
\newcommand{\seclab}[1]{\label{sec:#1}}

\newcommand{\figref}[1]{Figure~\ref{fig:#1}}
\newcommand{\figlab}[1]{\label{fig:#1}}

\newcommand{\tblref}[1]{Table~\ref{tbl:#1}}
\newcommand{\tbllab}[1]{\label{tbl:#1}}

\title{What Goes Where: Predicting Object Distributions from Above} 
\name{Connor Greenwell, Scott Workman, Nathan Jacobs}
\address{
(connor,scott,jacobs)@cs.uky.edu \\
Department of Computer Science, University of Kentucky, USA}

\begin{document}
%
\maketitle
\begin{abstract}

  
  In this work, we propose a cross-view learning approach, in which
  images captured from a ground-level view are used as weakly
  supervised annotations for interpreting overhead imagery.  The
  outcome is a convolutional neural network for overhead imagery that
  is capable of predicting the type and count of objects that are
  likely to be seen from a ground-level perspective.  We demonstrate
  our approach on a large dataset of geotagged ground-level and
  overhead imagery and find that our network captures semantically
  meaningful features, despite being trained without manual
  annotations.


\end{abstract}
\section{Introduction}
\seclab{intro}

The goal of remote sensing is to use imagery to obtain some
understanding of a particular location. Observations obtained from
satellites and aerial imaging have long been used to monitor the
Earth's surface. For example, to map land use, predict the weather,
understand urban infrastructure, and enable precision agriculture. A key
challenge is that obtaining labeled training data for new tasks can be
prohibitively expensive, especially if many manual
annotations are required. 

Recently, a significant amount of work has explored how deep learning
techniques can be applied to remote sensed data (see \cite{deepRS} for
a comprehensive review). We propose to use overhead imagery to
understand the type and quantity of objects one would expect to see at
a particular location.  Instead of acquiring manual annotations, we
consider labels inferred from nearby geotagged social media.
Specifically, we use an off-the-shelf object detector, applied to
ground-level imagery, to learn to interpret overhead imagery, a
special case of what we call cross-view semantic
transfer~\cite{zhai2017predicting}. Cross-view training approaches
have been applied to a variety of tasks, for example image
geolocalization~\cite{workman2015wide}, image-driven
mapping~\cite{workman2017beauty}, and constructing aural 
atlases~\cite{salem2018soundscape}.

In our approach, we train a convolutional neural network, operating on
overhead imagery, to predict the distribution of objects obtained from
co-located ground-level imagery.  We present results which demonstrate
how such an approach captures visual representations that are
geo-informative, despite being trained without manual annotations.

\section{Problem Statement}

\begin{figure*}
  \includegraphics[width=\linewidth]{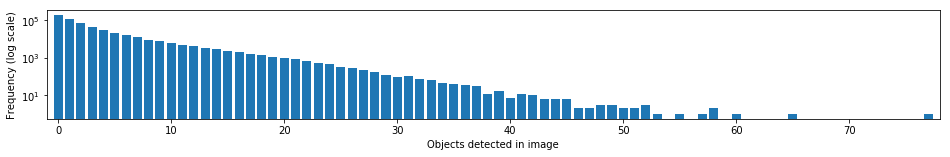}
  \includegraphics[width=\linewidth]{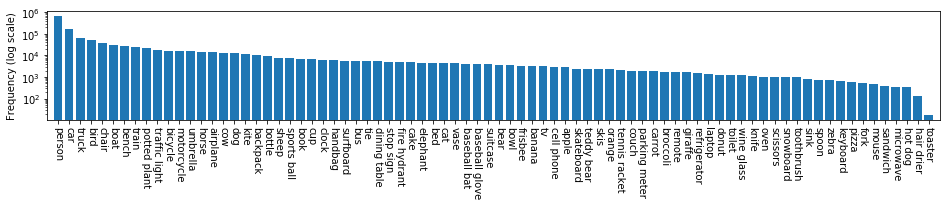}
  \caption{Object count histograms in the CVUSA dataset. (top) This
  histogram shows that most images contain very few objects. (bottom)
This histogram shows that person, car, and truck are the most
frequently detected object types.}
  \figlab{counts}
\end{figure*}

\begin{figure}[t]
  \centering
  \subcaptionbox{Person}{\includegraphics[width=.9\linewidth]{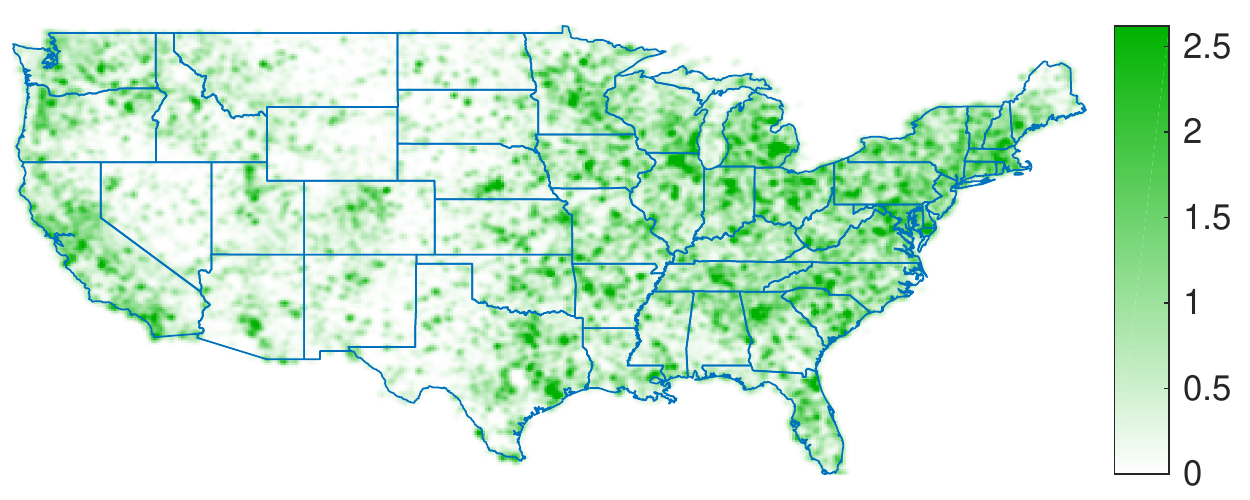}}
  \subcaptionbox{Train}{\includegraphics[width=.9\linewidth]{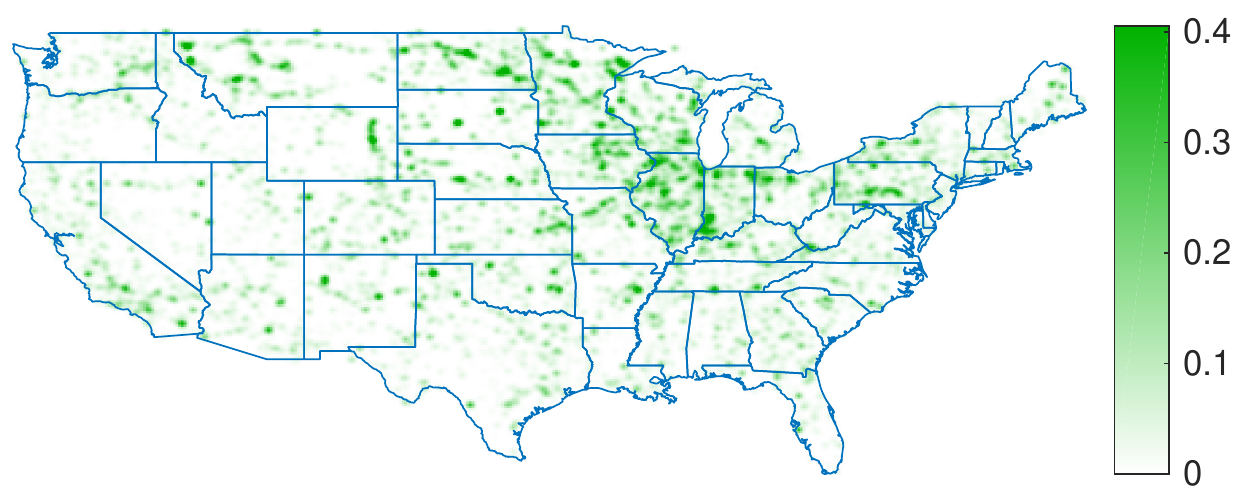}}
  \subcaptionbox{Truck}{\includegraphics[width=.9\linewidth]{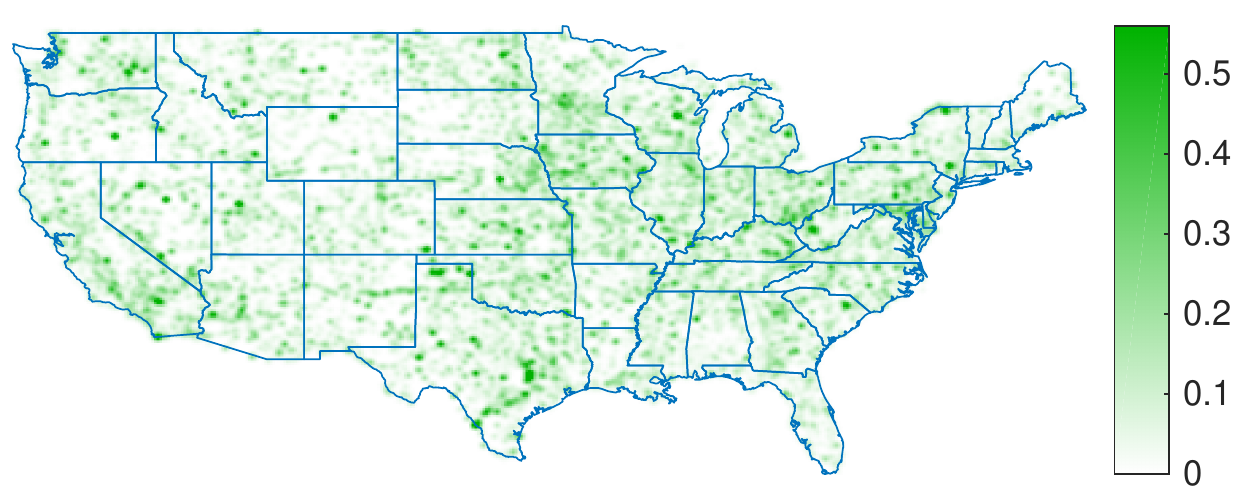}}
  \caption{A low resolution heatmap of object frequency generated using our baseline method (\secref{dataset}). Greener
(darker) locations mean that an image at that location will typically feature more of that type of object.}
  \figlab{dists}
\end{figure}

The goal of this work is to estimate the expected distribution of
objects for a given location, $P(O | L)$, where $O$ is a histogram of
objects and $L$ is a geographic location. Estimating this distribution
by directly conditioning on geographic location is challenging,
because it would require that the distribution essentially memorize
the entire Earth.  We attempt to overcome this challenge by
conditioning the distribution on the overhead image of a location.
This makes intuitive sense because it is often possible to infer the
type of objects that would be present at a particular location from
an overhead perspective. Therefore, we focus on a particular form of this
distribution, specifically, $P(O|S(L))$, where $S(L)$ is an overhead image,
perhaps captured from a satellite or an airplane, of the location,
$L$. In the remainder of this section, we describe how we constructed
a dataset to support learning such a conditional distribution.

\subsection{Dataset}\seclab{dataset}

To construct a suitable dataset for evaluating our proposed methods, we begin with the
Cross-View USA (CVUSA) dataset~\cite{workman2015wide}, which was
originally created to support training models for image
geolocalization. In consists of 1,588,655 geotagged ground-level
images, 551,851 of which are from Flickr, the remainder of which are
from Google Street View. While there are three overhead images for
each ground-level image, we only use the one with the highest
resolution. 

We use the Faster R-CNN ResNet 101~\cite{ren2015faster} detector
trained on the MS-COCO challenge dataset~\cite{lin2014microsoft} to
detect objects in the CVUSA Flickr images. The activations from the
final output layer are thresholded at $0.5$. Instances of each class
with score above the threshold are tallied up to form a histogram
describing the objects present in each image. This was
implemented in the TensorFlow Object Detection
API~\cite{huang2016speed}. 

In \figref{counts}, we show two histograms of object counts for ground-level images in the CVUSA dataset, which ranges from 0
to 78.  A majority of the images in the dataset have at least one objected detected. On average each image contains 2.63
objects, excluding images with zero detections.  The most frequently detected object category is person.  As a baseline
approach to mapping object distributions, \figref{dists} shows a simply a locally-weighted average of object counts from each
ground-level image.  Several interesting patterns emerge, such as the extensive rail network near Chicago and major truck
routes across the United States. While these patterns reflect our expectations, due to the sparsity of the imagery it is not
possible to construct a high-resolution map in this manner. 



\section{Learning to Predict Object Distributions}

In this section, we describe our approach, which we call {\em
WhatGoesWhere (WGW)}, for predicting the geospatial distribution of
ground-level objects. We use the cross-view learning framework, in
which we train a network to interpret overhead imagery by having it
predict features extracted from ground-level images. This allows us to
learn to extract useful features from overhead imagery without the
need for manual annotation. 

Our model for predicting ground-level object counts from an overhead image ({\em WGW-P}) is based on the ResNet50
architecture~\cite{he2016deep}. We appended two 2048D Dense-BatchNorm-LeakyReLU layers and a final 91D Dense layer. The
outputs of this final layer encode the parameters to a collection of 91 Poisson distributions over object counts, one
distribution per MS-COCO object category. We also train two additional
models: the first based on the Negative Binomial
distribution ({\em WGW-NB}) with two 91D output layers and the second based on the Gaussian distribution ({\em
WGW-G}) with two 91D output layers. The final output layers of each model are passed through a
softplus to ensure that the outputs are strictly greater than zero.

We initialize the ResNet50 portion of the model with weights trained
on the ImageNet task~\cite{russakovsky2015imagenet}, and the
subsequent Dense layers with Glorot Uniform random
noise~\cite{bengio2011deep}.  During training we minimize the mean 
negative log likelihood of the resulting distributions.  All models
were trained using the Nesterov-Adam optimizer with a learning rate of
$2e^{-5}$. 


\section{Evaluation}
\seclab{eval}

\begin{table}
    \centering
    \caption{Quantitative results comparing models with different loss
    functions. Higher is better.}
    \tbllab{log-likelihood}
    \begin{tabular}{l | l | c}
        Method & Distribution & Mean Log-Likelihood \\ \hline
        {\em WGW-P} & Poisson & $-0.1214$ \\
        {\em WGW-NB} & Neg.\ Binomial & $-0.1441$ \\
        {\em WGW-G} & Gaussian & $-0.3425$ \\
    \end{tabular}
\end{table}

We evaluated our {\em WGW} models quantitatively on a randomly selected subset
(25\%) of
the ground-level images in the CVUSA dataset, each with an object histogram 
and corresponding overhead image. For each overhead image, we
predict the parameters of a probability distribution over object
counts. Then, for each ground-level image, we measure the likelihood
of the empirical object counts under the predicted distribution.
\tblref{log-likelihood} shows the mean log-likelihood on the test set
of our different models. The model based on the Poisson distribution,
{\em WGW-P}, produces the largest log-likelihood.  Therefore, for all
remaining evaluation we focus on this model.


%

We qualitatively evaluate {\em WGW-P} using 
the dense overhead imagery in the San Francisco
database~\cite{workman2015geocnn} to generate fine-grained maps over
a large, diverse region. \figref{sf-poisson} visualizes the results of
this experiment as a heatmap of expected counts for a subset of
object classes. We observe that the model learns to discriminate using
visual cues found in overhead imagery and that the results appear to
be geographically consistent. For example, \figref{sf-poisson} (f)
shows that cars are most likely to be found in urban areas, while (b), (d), and (g) show that boats, surfboards, and birds are
all found around major bodies of water. These heatmaps are much higher resolution than those shown in \figref{dists}.

To further visualize what {\em WGW-P} has learned, we present the
overhead images from San Francisco which maximize the expected count
of several object categories in \figref{sf-most}. For example, the
overhead image for person is of a stadium, surfboard is of a beach,
and airplane is of an airport. 

In \figref{sf-clusters} we show the results of performing k-means
clustering ($k=10$) on the predicted Poisson distribution parameters
for this area.  As shown, this process results in 
visually coherent regions where one can expect to find objects in
similar quantities. For example, the red cluster appears to most
highly correlate with aquatic areas. 

\begin{figure}[t!]
  \centering
  \subcaptionbox{Person}{\includegraphics[width=0.45\linewidth]{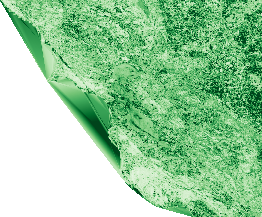}}
  \subcaptionbox{Boat}{\includegraphics[width=0.45\linewidth]{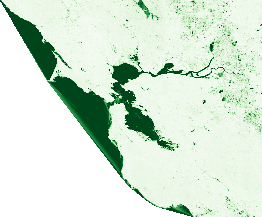}}
  \subcaptionbox{Train}{\includegraphics[width=0.45\linewidth]{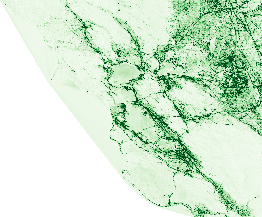}}
  \subcaptionbox{Surfboard}{\includegraphics[width=0.45\linewidth]{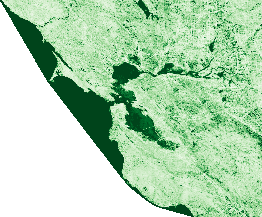}}
  \subcaptionbox{Truck}{\includegraphics[width=0.45\linewidth]{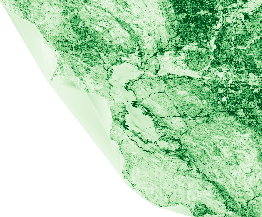}}
  \subcaptionbox{Car}{\includegraphics[width=0.45\linewidth]{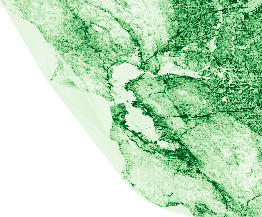}}
  \subcaptionbox{Bird}{\includegraphics[width=0.45\linewidth]{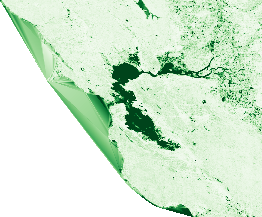}}
  \subcaptionbox{Airplane}{\includegraphics[width=0.45\linewidth]{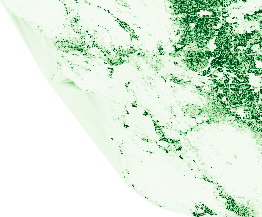}}

  \caption{A high resolution heatmap of object frequency generated using our {\em WGW-P} method.  Each map is scaled such that
the darkest (greenest) regions represent areas where we expect to see comparatively higher counts within an object category.
} 
  \figlab{sf-poisson}
\end{figure}


\begin{figure}[t]
  \centering 

  \subcaptionbox{Person}{\includegraphics[width=0.23\linewidth]{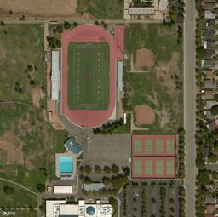}}
  \subcaptionbox{Train}{ \includegraphics[width=0.23\linewidth]{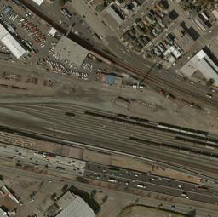}}
  \subcaptionbox{Truck}{ \includegraphics[width=0.23\linewidth]{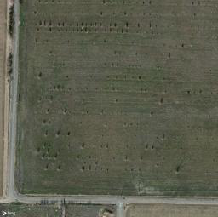}}
  \subcaptionbox{Boat}{  \includegraphics[width=0.23\linewidth]{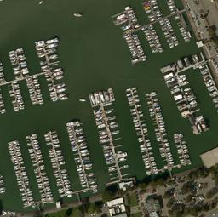}}

  \subcaptionbox{Surfboard}{\includegraphics[width=0.23\linewidth]{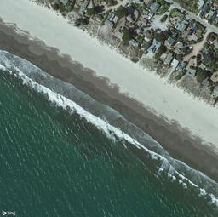}}
  \subcaptionbox{Car}{      \includegraphics[width=0.23\linewidth]{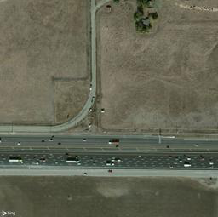}}
  \subcaptionbox{Bird}{     \includegraphics[width=0.23\linewidth]{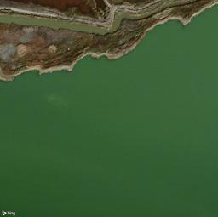}}
  \subcaptionbox{Airplane}{ \includegraphics[width=0.23\linewidth]{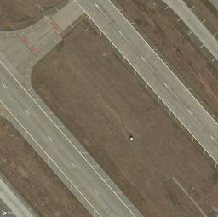}}

  \caption{Images that result in high expected counts for particular object categories, as estimated by our model. This figure
    shows that {\em WGW-P} learns to identify areas where large groups of the same object are often seen together. For
  example: (a) people at stadiums, (d) boats in marinas, and (h) airplanes at air strips.}
  \figlab{sf-most}
\end{figure}


\begin{figure}[t]
  \includegraphics[width=\linewidth]{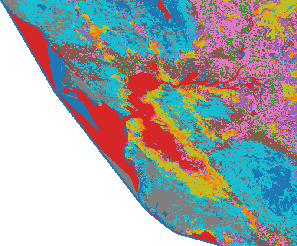}
  \caption{The k-means clustering over expected object counts for
  overhead imagery captured over San Francisco. Each location is colored based on
  which cluster to which it is assigned. The clusters appear to roughly correspond with terrain types in the area. For example,
red with the Ocean/Bay, green and orange with the urban areas that encircle the Bay, etc.}
  \figlab{sf-clusters}
\end{figure}

\section{Conclusion}

We proposed to use an off-the-shelf object detector, applied to
ground-level imagery, to learn to interpret overhead imagery, a
special case of what we call crossview semantic transfer. The 
idea was to use pairs of co-located overhead and ground-level images
and train a CNN to predict the distribution of objects in the
ground-level image using only the overhead image. We demonstrated how
this is able to capture rich and subtle differences between different
locations.  In some sense, what we learned is similar to land cover
and land use classification, the difference is that our approach does
not require us to commit to a particular set of classes in advance.
Instead, the mixture of object types that are likely to occur in an
area informs the types of representations we learn.  There are many
directions for future work, including applying this strategy to other
off-the-shelf methods for interpreting ground-level imagery and using
this as a pre-training strategy for a wide variety of overhead image
interpretation tasks.

\subsection*{Acknowledgments} 
We gratefully acknowledge the support of an NSF CAREER award
(IIS-1553116).

{
  \small
\bibliographystyle{IEEEbib}
\bibliography{biblio}

\begin{thebibliography}{10}

\bibitem{deepRS}
Xiao~Xiang Zhu, Devis Tuia, Lichao Mou, Gui{-}Song Xia, Liangpei Zhang, Feng
  Xu, and Friedrich Fraundorfer,
\newblock ``Deep learning in remote sensing: a review,''
\newblock {\em CoRR}, vol. abs/1710.03959, 2017.

\bibitem{zhai2017predicting}
Menghua Zhai, Zach Bessinger, Scott Workman, and Nathan Jacobs,
\newblock ``Predicting ground-level scene layout from aerial imagery,''
\newblock in {\em IEEE Conference on Computer Vision and Pattern Recognition
  (CVPR)}, 2017.

\bibitem{workman2015wide}
Scott Workman, Richard Souvenir, and Nathan Jacobs,
\newblock ``Wide-area image geolocalization with aerial reference imagery,''
\newblock in {\em IEEE International Conference on Computer Vision (ICCV)},
  2015.

\bibitem{workman2017beauty}
Scott Workman, Richard Souvenir, and Nathan Jacobs,
\newblock ``Understanding and mapping natural beauty,''
\newblock in {\em IEEE International Conference on Computer Vision (ICCV)},
  2017.

\bibitem{salem2018soundscape}
Tawfiq Salem, Menghua Zhai, Scott Workman, and Nathan Jacobs,
\newblock ``{A Multimodal Approach to Mapping Soundscapes},''
\newblock in {\em {IEEE International Geoscience and Remote Sensing Symposium
  (IGARSS)}}, 2018.

\bibitem{ren2015faster}
Shaoqing Ren, Kaiming He, Ross Girshick, and Jian Sun,
\newblock ``Faster r-cnn: Towards real-time object detection with region
  proposal networks,''
\newblock in {\em Advances in Neural Information Processing Systems (NIPS)},
  2015.

\bibitem{lin2014microsoft}
Tsung-Yi Lin, Michael Maire, Serge Belongie, James Hays, Pietro Perona, Deva
  Ramanan, Piotr Doll{\'a}r, and C~Lawrence Zitnick,
\newblock ``Microsoft coco: Common objects in context,''
\newblock in {\em European Conference on Computer Vision (ECCV)}, 2014.

\bibitem{huang2016speed}
Jonathan Huang, Vivek Rathod, Chen Sun, Menglong Zhu, Anoop Korattikara,
  Alireza Fathi, Ian Fischer, Zbigniew Wojna, Yang Song, Sergio Guadarrama,
  et~al.,
\newblock ``Speed/accuracy trade-offs for modern convolutional object
  detectors,''
\newblock {\em arXiv preprint arXiv:1611.10012}, 2016.

\bibitem{he2016deep}
Kaiming He, Xiangyu Zhang, Shaoqing Ren, and Jian Sun,
\newblock ``Deep residual learning for image recognition,''
\newblock in {\em IEEE Conference on Computer Vision and Pattern Recognition
  (CVPR)}, 2016.

\bibitem{russakovsky2015imagenet}
Olga Russakovsky et~al.,
\newblock ``Imagenet large scale visual recognition challenge,''
\newblock {\em International Journal of Computer Vision}, vol. 115, no. 3, pp.
  211--252, 2015.

\bibitem{bengio2011deep}
Yoshua Bengio et~al.,
\newblock ``Deep learners benefit more from out-of-distribution examples,''
\newblock in {\em International Conference on Artificial Intelligence and
  Statistics}, 2011.

\bibitem{workman2015geocnn}
Scott Workman and Nathan Jacobs,
\newblock ``On the location dependence of convolutional neural network
  features,''
\newblock in {\em IEEE/ISPRS Workshop: EARTHVISION: Looking From Above: When
  Earth Observation Meets Vision}, 2015.

\end{thebibliography}
}

\end{document}